\documentclass[letterpaper]{article} 
\usepackage{aaai25}  
\usepackage{times}  
\usepackage{helvet}  
\usepackage{courier}  
\usepackage[hyphens]{url}  
\usepackage{graphicx} 
\usepackage{natbib}  
\usepackage{caption} 
\usepackage{algorithm}
\usepackage{algorithmic}
\usepackage{amsmath}
\usepackage{amssymb}
\usepackage{amsfonts}
\usepackage{enumitem}
\usepackage{booktabs}
\usepackage{subcaption}
\usepackage{comment}

\usepackage{newfloat}
\usepackage{listings}
\DeclareCaptionStyle{ruled}{labelfont=normalfont,labelsep=colon,strut=off} 
\floatstyle{ruled}
\newfloat{listing}{tb}{lst}{}
\floatname{listing}{Listing}
\nocopyright

\title{Adaptive Visual Navigation Assistant in 3D RPGs}
\author{
    Kaijie Xu, Clark Verbrugge\\
}
\affiliations{
    Department of Computer Science, McGill University\\
    Montreal, Quebec, Canada\\
    kaijie.xu2@mail.mcgill.ca, clump@cs.mcgill.ca
}

\begin{document}

\maketitle

\begin{abstract}
In complex 3D game environments, players rely on visual affordances to spot map transition points. Efficient identification of such points is important to client-side auto-mapping, and provides an objective basis for evaluating map cue presentation. In this work, we formalize the task of detecting traversable Spatial Transition Points (STPs)-connectors between two sub regions-and selecting the singular Main STP (MSTP), the unique STP that lies on the designer-intended critical path toward the player’s current macro-objective, from a single game frame, proposing this as a new research focus. We introduce a two-stage deep-learning pipeline that first detects potential STPs using Faster R-CNN and then ranks them with a lightweight MSTP selector that fuses local and global visual features. Both stages benefit from parameter-efficient adapters, and we further introduce an optional retrieval-augmented fusion step. Our primary goal is to establish the feasibility of this problem and set baseline performance metrics. We validate our approach on a custom-built, diverse dataset collected from five Action RPG titles. Our experiments reveal a key trade-off: while full-network fine-tuning produces superior STP detection with sufficient data, adapter-only transfer is significantly more robust and effective in low-data scenarios and for the MSTP selection task. By defining this novel problem, providing a baseline pipeline and dataset, and offering initial insights into efficient model adaptation, we aim to contribute to future AI-driven navigation aids and data-informed level-design tools.
\end{abstract}

\begin{links}
   \link{Code}{https://github.com/Nortrom1213/VisualGuidance}
\end{links}

\section{Introduction}
\label{sec:introduction}

The design of game levels profoundly shapes player experience, directly impacting engagement, learning, and progression in virtual worlds \cite{yalcinakaya_material_matters_2024}. A key component of successful level design is the effective use of visual cues to guide players, particularly in complex 3D environments. Such guidance aims to enable intuitive navigation and reduce cognitive load, often by leveraging environmental affordances—properties suggesting action possibilities—to direct attention and movement \cite{gibson2014ecological, irshad2021wayfinding}. Designers use visual strategies like lighting, color, landmarks, and architecture to subtly orient players and indicate paths, fostering deeper immersion than explicit aids might allow \cite{dillman2018visual}.

Although visual guidance principles are well studied in design and HCI, most analyses are qualitative or center on player responses rather than the cues' intrinsic, measurable properties \cite{ixdf_games_user_research_qualitative}. Existing work probes perceptions \cite{boe2024role} or compares task completion under varied cues \cite{filen2024auditory}, but a systematic, computational account of what makes a visual cue effective for navigation remains largely open—highlighting an opportunity for data-driven methods to quantify the visual signatures of navigational cues.

\begin{figure*}[h!]
  \centering
  
  \begin{subfigure}[b]{0.48\textwidth}
    \centering
    \includegraphics[width=\textwidth]{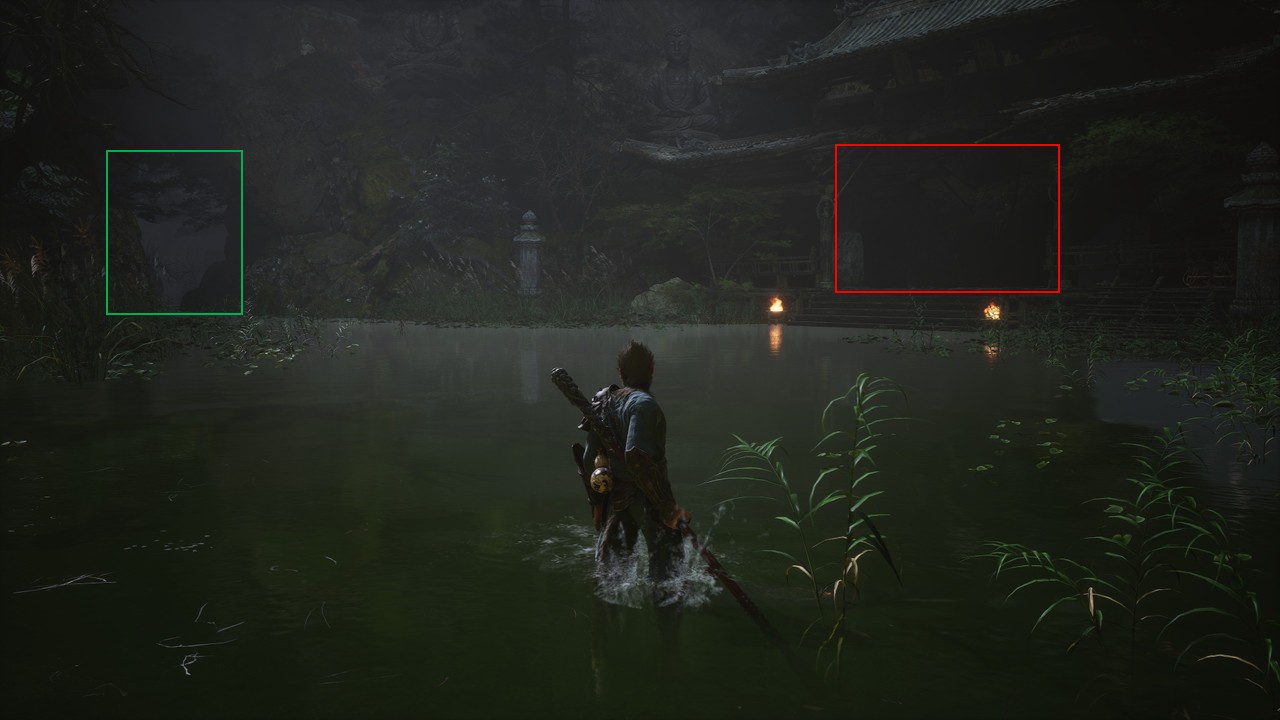} 
    \caption{Player assistance: Green box highlights a hidden optimal boss route that players often miss; red box marks the normal main route.}
    \label{fig:player_assist_example_sub}
  \end{subfigure}
  \hfill 
  \begin{subfigure}[b]{0.48\textwidth}
    \centering
    \includegraphics[width=\textwidth]{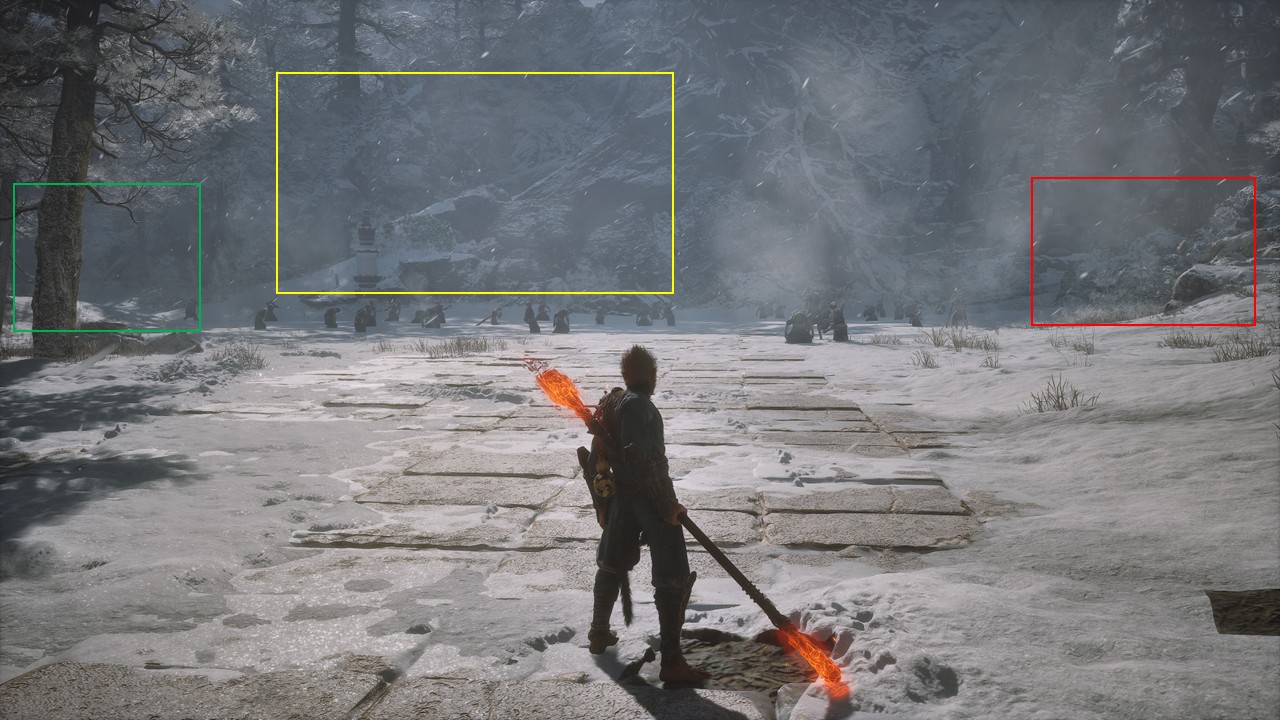} 
    \caption{Design evaluation: Green marks a secondary STP; red the designer-intended MSTP; yellow a deceptive, impassable ``air wall.''}
    \label{fig:designer_aid_example_sub}
  \end{subfigure}
  
  \caption{Examples of the system's application: (a) Assisting players by identifying easily missed STPs like hidden paths. (b) Aiding designers by evaluating visual guidance towards the intended MSTP versus potentially misleading alternatives.}
  \label{fig:application_examples} 
\end{figure*}

This paper investigates the formalization and automation of recognizing such visual cues—Spatial Transition Points (STPs), the passable links between map regions—and the Main STP (MSTP), the most prominent STP that represents the designer’s intended route to the player’s current objective, all from a single game frame. We propose this as a new research focus, hypothesizing that learnable visual metrics within game environments correlate with these navigational elements. Identifying such metrics presents significant potential for applications like (1) post-development navigation assistants that subtly highlight implicit cues for struggling players (Figure~\ref{fig:application_examples}a), and (2) pre-development design aids offering real-time feedback on visual clarity (Figure~\ref{fig:application_examples}b).

To explore this, we introduce a two-stage deep learning pipeline that analyzes visual data. The system first detects potential STPs and then identifies the MSTP, employing contemporary computer vision techniques and parameter-efficient adaptation methods. The primary aim of our work, validated on a newly custom-built, diverse dataset from five Action RPGs (\textit{Dark Souls I, II, III}, \textit{Elden Ring}, \textit{Black Myth: Wukong}), is to establish the initial feasibility and tractability of this novel problem by evaluating our baseline solution under varied data conditions. Our key contributions are:
\begin{itemize}[itemsep=0pt,leftmargin=*,nosep]
    \item The formulation of automated STP/MSTP recognition from visual game data as a new research problem, with a proposed two-stage pipeline as a baseline solution.
    \item A new, diverse, richly annotated multi-game dataset for visual navigation analysis, to be publicly released to foster research and comparative studies.
    \item Initial empirical validation confirming the pipeline efficacy, particularly in low-resource, cross-domain settings.
\end{itemize}

\section{Background}
\label{sec:background}

Our work builds upon several established areas of research. This section reviews relevant literature on visual analysis in game environments, computational methods for player, level analysis, and the application of deep learning in games.

\subsection{Visual Perception, Affordances, and Wayfinding}
\label{subsec:visual_perception_wayfinding}
The way players perceive and navigate virtual environments is deeply influenced by the visual information presented to them. Game environments often tell stories and guide players through their spatial design, a practice termed environmental storytelling \cite{fernandez2011game}. This relates to Gibson's (\citeyear{gibson2014ecological}) ecological view of affordances, where environmental forms themselves suggest how one might interact with them. In games, these affordances are often communicated through carefully designed visual cues that signal what can be interacted with, where to look, or where to go \cite{dillman2018visual}. Effective wayfinding in games relies on players' ability to form cognitive maps of the environment, often aided by landmarks, distinct paths, and spatial organization \cite{lynch2023image} in urban design and applicable to virtual spaces \cite{yalcinakaya_material_matters_2024}. HCI research for games investigates how these elements contribute to spatial knowledge and navigation efficiency, particularly in immersive virtual reality environments that emphasize embodied interaction \cite{lin2024design, marcus2025hci}. Our focus on STPs and MSTPs is an attempt to computationally model the perception of critical navigational affordances based on their visual presentation.

\subsection{Player Guidance and Level Analysis}
\label{subsec:computational_guidance_level_analysis}
Beyond manual design principles, computational methods are increasingly used to analyze and enhance player navigation within game levels. Player modeling, for instance, aims to capture player states computationally, enabling adaptive systems that can modify guidance strategies \cite{yannakakis2018artificial, hare2022player}. Such models might analyze gameplay data to identify navigational challenges and inform personalized support. Separately, automated playtesting uses AI agents to evaluate level traversability or to identify gameplay imbalances by training agents to navigate based on environmental stimuli \cite{miller2024automated}. Other research applies graph-based or metric-driven analyses to evaluate the topological structure of game levels, which can indirectly relate to their navigability \cite{11114135, omidshafiei2020navigating, naying2023computational}. Our research extends these efforts by focusing on the direct visual analysis of game scenes for explicit navigational markers, distinct from approaches on player telemetry or abstract spatial graphs.

\subsection{Deep Learning for Visual Scene Understanding}
\label{subsec:deep_learning_visual_scene_games}
Deep learning techniques, particularly Convolutional Neural Networks (CNNs), have significantly enhanced machines' abilities to parse complex visual information. These methods are applied to tasks such as object detection, semantic segmentation, and overall scene understanding \cite{betsas2025deep, qi2023application}. In game research, deep learning has been used for diverse applications, from procedural content generation to controlling non-player characters and analyzing player affect \cite{liu2021deep, mehta2025role}. Object detection systems such as Faster R-CNN or YOLO have also been adapted to identify elements in games; however, their effectiveness often depends on specialized fine-tuning due to diverse artistic styles and visual complexities \cite{jung2021improving, hyde2023using}. Furthermore, understanding dynamic scenes and player attention within them is an active area of research, with models aiming to predict human players' focus \cite{yazdani2025computational}. The challenge of adapting models trained on one visual domain (e.g., real-world images or a specific game) to another (e.g., a different game with a distinct art style) is addressed by domain adaptation techniques, which are crucial for creating broadly applicable visual recognition systems for games \cite{csurka2017domain, patel2015visual}. Our use of deep learning for detecting STPs/MSTPs and employing adapter modules for fine-tuning aligns with these trends, aiming to create a robust visual analysis tool for diverse game environments.

\section{Methodology}

In this section, we formalize the STP/MSTP detection task and present our two-stage pipeline—STP detection, MSTP selection, and optional retrieval-augmented fusion (RAF).

\subsection{Problem Definition}
A \textit{Spatial Transition Point (STP)} is any traversable doorway, ladder, corridor, or passage that links two distinct map regions. A \textit{Main Spatial Transition Point (MSTP)} is the unique STP on the designer-defined critical path toward the current macro-objective (inferred here as the next boss or primary level goal). We restrict our study to ARPGs with a clearly defined target (e.g., a boss encounter) in each single level for simplicity. The objective of this work is to develop an automated pipeline that, given a single input frame, (i) detects all STPs and (ii) identifies the single MSTP among them.

\subsection{Pipeline Overview}

\begin{figure}[!t]
  \centering
  \includegraphics[width=.5\textwidth]{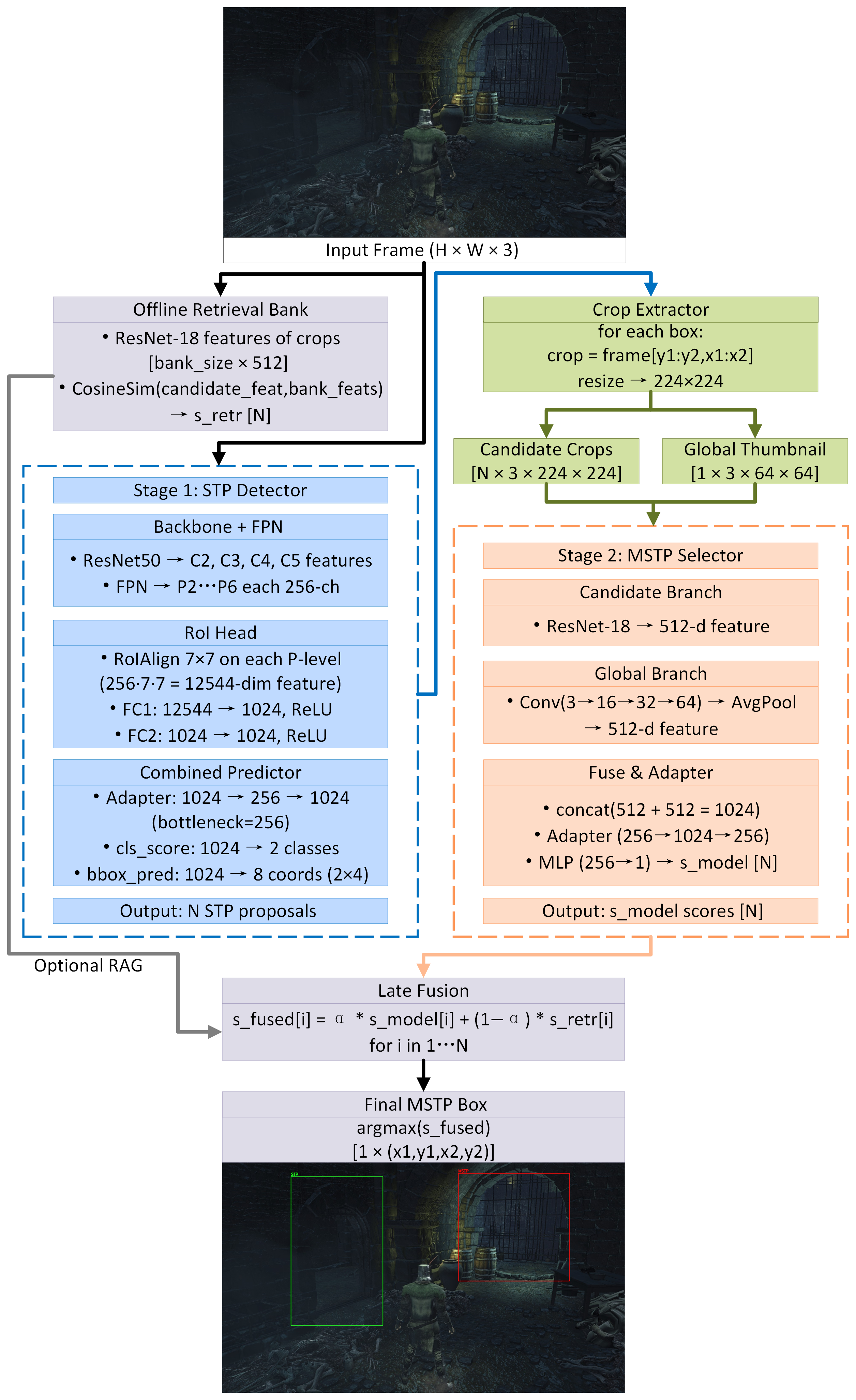}
  \caption{Top: example input frame. Center: the two‐stage pipeline. (1) Offline Retrieval Bank holds pre‐computed ResNet18 embeddings of annotated STP/MSTP crops. (2) STP Detector (Faster R-CNN + Adapter-augmented head) outputs $N$ candidate boxes. (3) For each box, we extract a $224\times224$ local crop and a $64\times64$ global thumbnail. (4) MSTP Selector fuses 512-d local and 512-d global features via an Adapter bottleneck and a two-layer MLP, yielding scores $s_i^{\mathrm{sel}}$. (5) We compute retrieval scores $s_i^{\mathrm{ret}}$ by cosine similarity against the bank, then form 
  $s_i^{\mathrm{final}}=\alpha\,s_i^{\mathrm{sel}}+(1-\alpha)\,s_i^{\mathrm{ret}}$. 
  The box with highest $s_i^{\mathrm{final}}$ is returned as the MSTP.}
  \label{fig:overview}
\end{figure}

We proceed in three logical blocks shown in Figure~\ref{fig:overview}:  
\begin{enumerate}[itemsep=0pt,leftmargin=*]
  \item \textbf{STP Detection:} a parameter‐efficient Adapter head fine-tunes Faster R-CNN to propose spatial transition points.
  \item \textbf{MSTP Selection:} a lightweight selector network combines local patch features and global image features (both processed through dedicated network branches, with their fusion augmented by the same bottleneck Adapter mechanism to rank each STP proposal.
  \item \textbf{Retrieval‐Augmented Fusion (optional):} we combine selector scores with cosine‐based retrieval scores from an offline feature bank, generating the final MSTP choice.
\end{enumerate}

We adopt this two-stage approach rather than direct simultaneous classification because identical physical structures or visual cues may represent different navigation significance depending on their spatial relationships, viewpoint, or surrounding visual context.

\subsection{Stage 1: STP Detection}

We adopt Faster R-CNN~\cite{ren2015faster} with a ResNet50-FPN backbone.
Given an RGB frame $\mathbf I$, the detector predicts $K$ bounding boxes
$\{\mathbf b_i\}_{i=1}^{K}$ and confidence scores
$\{s_i^{\mathrm{det}}\}_{i=1}^{K}$.  
The network is trained with standard Region Proposal Network (RPN) + classification + regression losses.

\subsection{Stage 2: MSTP Selection With Global Context}

For each detector proposal $\mathbf b_i$ we crop a
$224\times224$ local patch and pass it through a ResNet18 branch,
producing a feature vector $\mathbf{f}_i^{\mathrm{loc}} \in \mathbb{R}^{512}$.
In parallel, a $64\times64$ thumbnail of the whole frame feeds a lightweight CNN to obtain a global feature vector $\mathbf{f}^{\mathrm{glob}} \in \mathbb{R}^{512}$. This CNN, whose architecture is detailed in Figure~\ref{fig:overview} (Global Branch), transforms the thumbnail into a global feature vector $\mathbf{f}^{\mathrm{glob}} \in \mathbb{R}^{512}$ followed by ReLU activation.
The local feature vector $\mathbf f_i^{\mathrm{loc}}$ and the global feature vector $\mathbf f^{\mathrm{glob}}$ are concatenated, forming a 1024-dimensional vector $[\mathbf{f}_i^{\mathrm{loc}}; \mathbf{f}^{\mathrm{glob}}]$. This combined vector is then processed by an Adapter module with a bottleneck dimension of $r=256$, resulting in the refined feature vector $\mathbf f_i \in\mathbb R^{1024}$:
\[
\mathbf f_i = \mathrm{Adapter}\bigl([\mathbf f_i^{\mathrm{loc}}; \mathbf f^{\mathrm{glob}}]\bigr).
\]
The resulting vector $\mathbf f_i$ is then passed through a two-layer Multi-Layer Perceptron (MLP) to output a scalar score $s_i^{\mathrm{sel}}$. This MLP consists of a linear layer projecting from 1024 to 256 dimensions followed by a ReLU activation, and a final linear layer projecting to a single scalar score. Cross-entropy loss, applied to the scores of all candidate proposals, enforces the ground-truth MSTP to rank highest.

ResNet50-FPN was chosen for its strong performance in object detection tasks, while a lighter ResNet18 was selected for local feature extraction to maintain efficiency in the MSTP selection stage. The 64×64 thumbnail provides a global context without significant computational overhead.

\subsection{Parameter-Efficient Adapter Fine-tuning}

To enable fast adaptation to a new game with limited images, we insert a bottleneck Adapter following Houlsby (\citeyear{houlsby2019parameter}):
\begin{equation}
\label{eq:adapter}
\mathbf{x}' = \mathbf{x} + \mathbf W_\text{up}\bigl(\sigma(\mathbf W_\text{down}\,\mathbf{x})\bigr),
\end{equation}
where $\mathbf W_\text{down}\in\mathbb R^{r\times d}$, $\mathbf W_\text{up}\in\mathbb
R^{d\times r}$, $r \!\ll\! d$. Here, $\sigma$ represents the ReLU activation function. During normal training, the adapter parameters are initialized to zero, reducing Eq.~\eqref{eq:adapter} to the identity mapping without affecting capacity. For adapter fine-tuning, we freeze the backbone and optimize only $\{\mathbf W_\text{down},\mathbf W_\text{up}\}$ and the prediction heads.

\subsection{Retrieval-Augmented Late Fusion}
\label{sec:method_rag}

To further boost cross-game robustness, we build an \emph{offline
feature bank} $\mathcal B=\{(\mathbf f_j, \ell_j)\}_{j=1}^{M}$, where each $\mathbf f_j \in \mathbb{R}^{512}$ is a ResNet18 embedding of an annotated region, and $\ell_j$ is its corresponding class label (STP or MSTP) and source game information. This bank stores embeddings of STP/MSTP regions from various titles.  At inference, we:

\begin{enumerate}[itemsep=0pt,leftmargin=*]
  \item extract a feature $\hat{\mathbf f}_i$ for each candidate crop via the same ResNet18 branch,
  \item compute a retrieval score
        $s_i^{\mathrm{ret}}=\max_{(\mathbf f_j,\ell_j)\in\mathcal B}
        \cos(\hat{\mathbf f}_i,\mathbf f_j)$,
  \item fuse scores
        $
        s_i^{\mathrm{final}} = \alpha\,s_i^{\mathrm{sel}}
                           + (1-\alpha)\,s_i^{\mathrm{ret}},
        \quad\alpha\in[0,1],
        $
  \item pick the candidate with the largest $s_i^{\mathrm{final}}$ as the predicted MSTP.
\end{enumerate}

This late-fusion design requires no additional training and empirically
improves recall on unseen maps.

\section{Data Collection and Annotation}

In this section, we describe how we assembled our multi‐game ARPG screenshot dataset, labeled all STPs, and identified the MSTP. A full two-stage protocol with expert review is planned for future releases.

\subsection{Dataset Construction}
We manually constructed a dataset from four commercially released third-person ARPGs (\textit{Dark Souls I}, \textit{II}, \textit{III}, and \textit{Elden Ring}), chosen for their intricate, nonlinear level architectures and clearly defined macro-objectives. In addition, to verify cross-franchise generalization, we captured some frames from \textit{Black Myth: Wukong}, which—like the Souls series—is a third-person ARPG with explicit end-level targets that facilitate MSTP annotation.  BMW’s use of photogrammetric 3D scanning produces highly realistic, ``real-world'' visuals; while this dramatically enhances immersion, it also sometimes leads to spurious visual cues that can mislead naive transition-point detectors. All screenshots were taken at predetermined camera positions to ensure uniform coverage of potential transition areas. We disabled all in-game UI and held the player avatar in a fixed pose so that only the underlying environment informed our models. Each decision point was captured exactly once, ensuring that no single navigational junction appears in more than one frame across the dataset.

\subsection{Annotation Protocol}
\label{sec:annotation_protocol}
Each frame was manually annotated by an expert player (one of the authors) who drew polygonal bounding boxes around every STP and marked the single MSTP according to the hierarchy described below. To maximize consistency, the annotator followed guidelines derived from established player conventions and community-verified walkthroughs.

\paragraph{STP Definition and Criteria}
An STP is defined as a visually distinct, traversable region that connects two separate map sub-regions. For annotation, an STP must have:
\begin{itemize}[itemsep=0pt,leftmargin=*,nosep]
    \item \textbf{Clear Spatial Separation:} Evident environmental changes (in texture, lighting, geometry, or theme) signaling a transition.
    \item \textbf{Explicit Visual Cues:} Presence of unambiguous indicators such as arches, doors, stairways, or other common navigational affordances.
    \item \textbf{Functional Navigational Role:} The region must serve a recognizable transitional purpose aligned with game progression or established level structure.
\end{itemize}

\paragraph{MSTP Selection Hierarchy}
The unique MSTP within a frame was identified from the annotated STPs using a hierarchical filtering process:
\begin{enumerate}[itemsep=0pt,leftmargin=*,nosep]
    \item \textbf{Navigational Continuity:} Verification of a direct, standard traversable path.
    \item \textbf{Path Efficiency:} Prioritization of the shortest verified route to the primary level objective, confirmed via in-game path tracing.
    \item \textbf{Route Connectivity:} Preference for STPs that link multiple alternative routes or offer greater navigational flexibility, if ambiguity remained.
    \item \textbf{Designer Intent Confirmation:} In rare, highly ambiguous cases, consultation of official game documentation or expert design knowledge.
\end{enumerate}

\paragraph{Handling Ambiguities}
Annotation ambiguities were systematically addressed: Overlapping STPs were merged if structurally similar, otherwise annotated separately with documented overlap. Regions with unclear boundaries received an uncertainty rating and explanatory notes. If multiple STPs were viable MSTP candidates, the above hierarchy was strictly applied to select one. STPs accessible only via non-standard movement mechanics (e.g., obscure sequence breaks) were generally excluded from MSTP candidacy unless critical to intended progression and supported by design documentation.

\subsection{Final Dataset Summary}

In total, our annotated corpus comprises 699 frames: 59 from \textit{Dark Souls I}, 100 from \textit{Dark Souls II}, 184 from \textit{Dark Souls III}, 230 from \textit{Elden Ring}, and 126 from \textit{Black Myth: Wukong}.  For \textit{Dark Souls II} and \textit{Black Myth: Wukong}, we adopt a 20\%/80\% train/test split to assess performance under limited‐data conditions; for the remaining three titles, we use an 80\%/20\% split.  We will publicly release this entire dataset under an open‐access license and maintain an online repository with ongoing updates and expansions.  

\section{Experiments and Results}

To evaluate the effectiveness and generalizability of our proposed two-stage pipeline, we perform comprehensive experiments using three distinct evaluation strategies:

\begin{itemize}[leftmargin=*]
    \item \textbf{Original dataset:} Comprising Dark Souls I, Dark Souls III, and Elden Ring (\textit{DS I, DS III, ER}), split randomly into 80\% training and 20\% testing sets. This assesses model efficacy under standard training conditions.
    \item \textbf{Novel dataset (DS II):} Only 20\% of samples used as a training set, to assess model adaptability and generalization performance under limited-data conditions.
    \item \textbf{Novel dataset (BMW):} Comprising Black Myth: Wukong (\textit{BMW}), 126 frames with a 20\%/80\% train/test split, used to validate cross-game transferability on a realistic, third-person ARPG outside the Souls lineage.
\end{itemize}

We explore the following model variants:

\begin{enumerate}[label=\textbf{\Alph*.},leftmargin=*]
    \item Full training on the original dataset (\textbf{Full Version})
    \item Adapter-only on the original dataset (\textbf{Adapter Version})
    \item Continuing full training on the novel dataset after full training on original dataset (\textbf{Continue on Full Version})
    \item Adapter-only fine-tuning on the novel dataset after full training on original dataset (\textbf{Adapter on Full Version})
    \item Full training solely on the novel dataset (\textbf{New Full})
    \item Adapter-only training solely on the novel dataset (\textbf{New Adapter Version})
\end{enumerate}

We evaluate STP Detection (\textbf{Model 1}) and MSTP Selection (\textbf{Model 2}) across these datasets using the following standard metrics. Detailed results are in Tables~\ref{tab:model1_results} and \ref{tab:model2_results}.

\paragraph{Evaluation Metrics}
We assess Model 1 using standard object detection metrics:
\begin{itemize}[itemsep=0pt,leftmargin=*,nosep]
    \item \textbf{mAP@IoU$>$0.5 (mean Average Precision):} Measures detection quality, balancing precision and recall at 0.5 IoU (Intersection over Union) threshold. Higher is better.
    \item \textbf{Mean IoU:} Assesses localization accuracy by averaging the IoU of correctly detected STPs with their ground-truth boxes. Higher values indicate better bounding box fit.
    \item \textbf{Recall:} The proportion of all STPs that the model successfully identifies. Higher values mean fewer missed STPs.
    \item \textbf{Composite Score:} For Model 1 training, this average of the metrics guides checkpoint selection, balancing detection quality and completeness.
\end{itemize}
For Model 2 (MSTP Selection), a classification task of choosing the correct MSTP from candidates, we report:
\begin{itemize}[itemsep=0pt,leftmargin=*,nosep]
    \item \textbf{Accuracy:} The percentage of times the model correctly identifies the ground-truth MSTP. Given our setup (one true MSTP per candidate set), this directly reflects correct navigational choice. 
\end{itemize}
For all experiments, the training split was further divided into 80\%/20\% train/validation subsets (the held-out test split was never used for tuning). Checkpoints were selected based on the highest composite validation score (mAP@0.5, mean IoU, and recall) for Model~1 and validation accuracy for Model~2, with early stopping applied.

\paragraph{Retrieval‐Augmented Late Fusion}

To further enhance robustness—especially under domain shift to DS II and BMW—we assemble our offline feature bank $\mathcal B$ from three complementary sources:

\begin{itemize}[itemsep=0pt,leftmargin=*]
  \item \textbf{Cross‐game core bank:} For each title in the Original dataset (DS I, DS III, ER), we extract all annotated STP/MSTP region embeddings via the ResNet18 branch, compute a quality score (L2 norm plus 0.5×std), sort descending, and retain the top 100 per title.
  \item \textbf{DS II support bank:} We append \emph{all} STP/MSTP embeddings from the DS II training split.
  \item \textbf{BMW support bank:} Likewise, we include \emph{all} STP/MSTP embeddings from the BMW training split.
\end{itemize}

At test time, the retrieval score $s_i^{\mathrm{ret}}$ and final fused score $s_i^{\mathrm{final}}$ (using $\alpha=0.8$) are computed for each candidate crop according to the procedure detailed in the previous section. The candidate with the highest $s_i^{\mathrm{final}}$ is selected as the MSTP.

Our hybrid feature bank, merging a cross-game core with specific support for DS II and BMW, improved performance on these unseen games without further model training. Although RAF's current boost to scores is modest (Table~\ref{tab:model2_results}), its key advantage is being a training-free component. We see RAF as a promising avenue for systems that learn on the fly; in practice, the bank could be updated with new features from errors or new game areas, allowing continuous adaptation. For this study, however, we used a fixed bank to fairly test RAF's baseline contribution and its role as an alternative to adapter fine-tuning.

\paragraph{Training details} 

Model~1 (STP detection) used Faster~R-CNN with ResNet50-FPN, SGD (lr=$5\times10^{-3}$, momentum=0.9, weight decay=$5\times10^{-4}$), batch size 4, and horizontal flip augmentation ($p=0.5$). Model~2 used a ResNet-18 local branch and a lightweight global CNN, SGD (lr=$1\times10^{-3}$, momentum=0.9, weight decay=$5\times10^{-4}$), batch size 4, 500 epochs, and adapter bottleneck $r=256$; the backbone was frozen except for adapters and the prediction head. Local crops were $224{\times}224$ and global thumbnails $64{\times}64$. Unless noted otherwise, the random seed was fixed at 42, with multi-seeds starting from 0.

\subsection{STP Detection Results (Model 1)}
\subsubsection{Learning Curves}

\begin{figure*}[htbp]
  \centering
  \begin{subfigure}{\textwidth}
    \includegraphics[width=.85\textwidth]{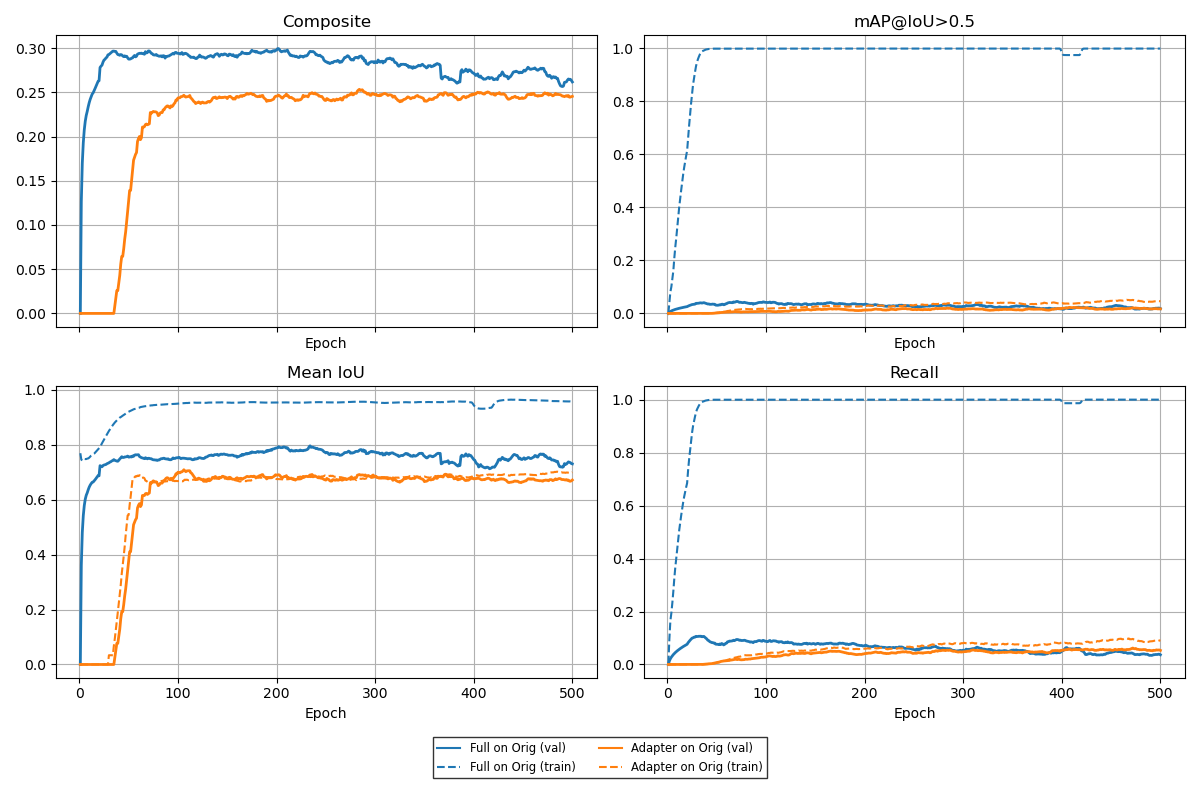}
    \caption{Validation on the original dataset (\textit{DS\,I, III, ER}).}
    \label{fig:m1_val_ori}
  \end{subfigure}
  \hfill
  \begin{subfigure}{\textwidth}
    \includegraphics[width=.85\textwidth]{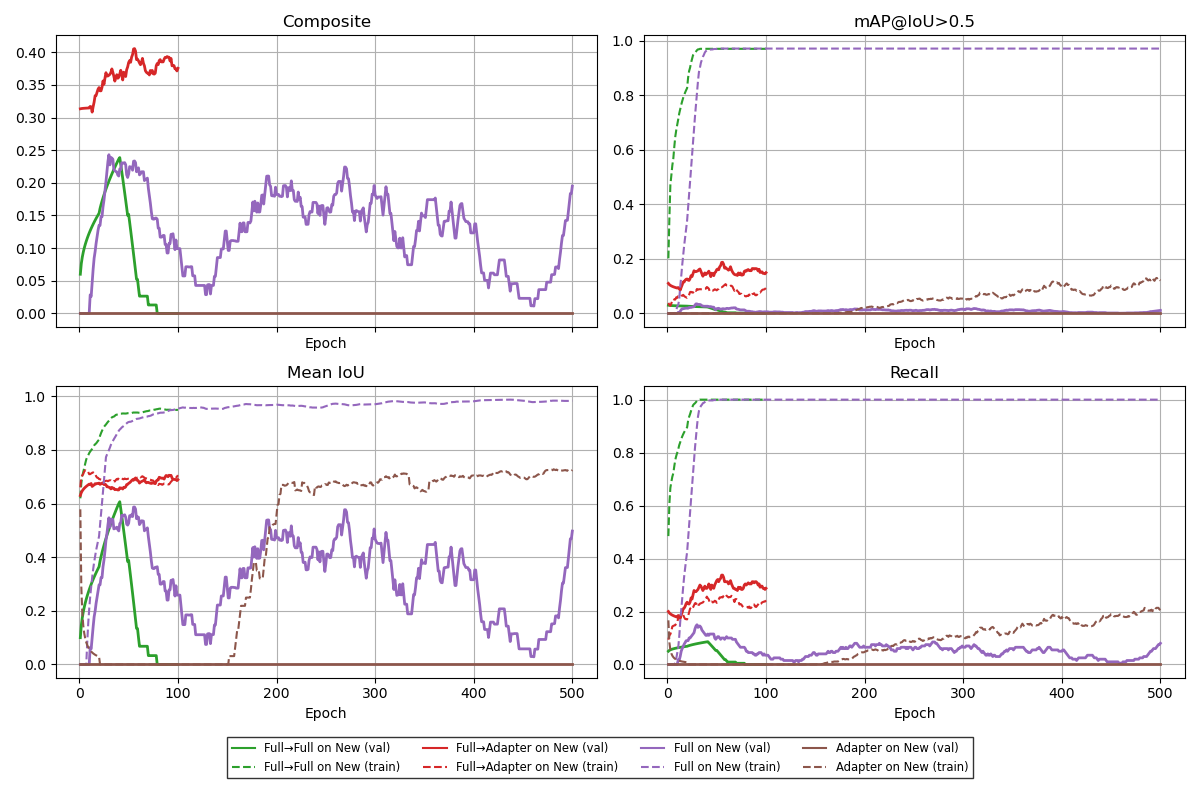}
    \caption{Validation on the limited‐data DS\,II split (20\% train / 80\% val).}
    \label{fig:m1_val_new}
  \end{subfigure}
  \caption{Model 1 (STP detection) training and validation curves.  For each setting we plot the \emph{validation} Composite score, mAP@IoU$>$0.5, Mean IoU and Recall (solid lines), together with the corresponding \emph{training} curves (dashed).  In (a) we compare Full vs.\ Adapter on the rich, original dataset; in (b) we compare four variants on the scarce DS II split.}
  \label{fig:model1_validation}
\end{figure*}

Figure~\ref{fig:model1_validation} presents training and validation curves for Model 1. Detecting STPs poses unique challenges: they are functionally defined, visually diverse, and often subtle, unlike typical objects in standard computer vision benchmarks. Our dataset is also modest in scale compared to large CV datasets. Thus, while absolute mAP and Recall values may seem low, our focus is on the relative performance of different training strategies and their generalization capabilities, which these curves clearly illustrate.

On the original dataset (Figure~\ref{fig:m1_val_ori}), the Full-network model consistently surpasses the Adapter-only variant. The gap between training and validation curves for the Full model indicates some overfitting, yet its validation performance is stable and superior, confirming the need for end-to-end fine-tuning with sufficient data.

In contrast, on the limited-data DS~II split (Figure~\ref{fig:m1_val_new}), models trained from scratch (``Full on New'' and ``Adapter on New'') show near-zero mAP and Recall. This failure to learn is expected given the extreme data scarcity (20 training images from DS~II), highlighting the difficulty of learning robust features without prior knowledge. Continuing full fine-tuning (``Full$\rightarrow$Full on New'') offers marginal gains. However, adapter-only transfer (``Full$\rightarrow$Adapter on New'') achieves higher performance across all metrics, showing that parameter-efficient adapters best preserve and transfer cross-game knowledge under severe data constraints.

Taken together, these curves confirm our core finding for STP detection: full-network training is essential when data are plentiful, but under extreme low-data conditions, adapter-only fine-tuning offers the most robust cross-game transfer. For clarity, we focus our curve discussion on Model 1, as Model 2 and the BMW dataset experiments show similar relative orderings, but with smaller absolute differences.

\begin{table}[t]
\centering
\resizebox{\linewidth}{!}{
\begin{tabular}{lcccc}
\toprule
Model Version            & Test Set     & mAP@IoU$>$0.5 (\%) & Mean IoU (\%) & Recall (\%) \\
\midrule
\multicolumn{5}{l}{\textit{Original Dataset (DS I, DS III, ER)}}\\
Full Version             & Original     & \textbf{22.04}& \textbf{72.05}& \textbf{36.31}        \\
Adapter Version          & Original     & 0.04                & 59.99         & 10.83        \\
\midrule
\multicolumn{5}{l}{\textit{Novel Dataset DS II}}\\
Full Version             & DS II        & 16.87               & 71.62         & 29.20        \\
Adapter Version          & DS II        & 0.05                & 60.34         & 8.03         \\
Continue on Full         & DS II        & 9.94                & 72.03         & 25.55        \\
Adapter on Full          & DS II        & \textbf{18.45}& \textbf{73.10}& \textbf{32.40}        \\
New Full Version         & DS II        & 3.43                & 71.03         & 13.87        \\
New Adapter Version      & DS II        & 0.07                & 58.97         & 20.44        \\
\midrule
\multicolumn{5}{l}{\textit{Novel Dataset BMW}}\\
Full Version             & BMW          & 7.64                & 63.95         & 13.58        \\
Adapter Version          & BMW          & 0.14                & 57.80         & 13.58        \\
Continue on Full         & BMW          & 8.34                & 61.89         & 35.80        \\
Adapter on Full          & BMW          & \textbf{9.50} & \textbf{64.50} & \textbf{38.20}        \\
New Full Version         & BMW          & 4.51                & 61.10         & 29.01        \\
New Adapter Version      & BMW          & 0.00                & 54.13         & 0.62         \\
\bottomrule
\end{tabular}}
\caption{Model 1 (STP Detection) Results. Bold values are the best results across models in each dataset category.}
\label{tab:model1_results}
\end{table}

\subsubsection{Analysis of STP Detection (Model 1)}

We did not include generic object-detection baselines because such models achieved a Mean IoU below 10\% in our experiments, providing virtually no usable information for this task and would not offer a meaningful comparison.   On the \textbf{original dataset}, the end‐to‐end fine‐tuned detector (Full Version) achieves its highest performance (mAP 22.04\%, IoU 72.05\%, Recall 36.31\%), while the Adapter‐only variant essentially fails (mAP 0.04\%, Recall 10.83\%).  This large gap confirms that adapters alone lack sufficient capacity for accurate STP localization when abundant data are available. 

Under the limited‐data \textbf{DS II} split, all models trained \emph{from scratch} on DS II collapse (New Full: mAP 3.43\%, Recall 13.87\%; New Adapter: mAP 0.07\%, Recall 20.44\%).  Continuing full fine‐tuning (Continue on Full) recovers some performance (mAP 9.94\%, Recall 25.55\%) but still lags behind zero‐shot transfer (Full Version: mAP 16.87\%, Recall 29.20\%).  Notably, adapter‐only transfer (Full→Adapter on New) now surpasses zero‐shot full transfer (mAP 18.45\% vs.\ 16.87\%, Recall 32.40\% vs.\ 29.20\%), confirming that parameter‐efficient adapters can not only preserve but even improve cross‐game detection under extreme scarcity.

On the \textbf{BMW} test split, a similar pattern emerges: scratch‐trained models remain weak (New Full: mAP 4.51\%, Recall 29.01\%; New Adapter: mAP 0.00\%, Recall 0.62\%), and zero‐shot full transfer yields moderate scores (mAP 7.64\%, Recall 13.58\%).  Continue‐on‐Full again boosts Recall (mAP 8.34\%, Recall 35.80\%), but adapter‐only transfer (Full→Adapter on New) now achieves the best (mAP 9.50\%, Recall 38.20\%), outpacing both zero‐shot and full fine‐tuning. This reinforces our core finding: under severe data scarcity—even in brand‐new domain—the adapter‐only strategy delivers the most robust cross‐game STP detection.

\subsection{MSTP Selection Results (Model 2)}

\begin{table}[t]
\centering
\resizebox{\linewidth}{!}{
\begin{tabular}{lcccc}
\toprule
Model Version            & Test Set & Accuracy (\%) & +RAF Accuracy (\%) \\
\midrule
\multicolumn{4}{l}{\it Original Dataset (DS I, DS III, ER)}\\
Full Version             & Original & 77.89         & 80.00              \\
Adapter Version          & Original & \textbf{80.00}         & 80.00             \\
\midrule
\multicolumn{4}{l}{\it Novel Dataset DS II}\\
Full Version             & DS II    & 66.25         & 72.50              \\
Adapter Version          & DS II    & 73.75         & 73.75              \\
Continue on Full Version & DS II    & 73.75         & --                 \\
Adapter on Full Version  & DS II    & 67.50         & --                 \\
New Full Version         & DS II    & 72.50         & --                 \\
New Adapter Version      & DS II    & \textbf{75.00}         & --                 \\
\midrule
\multicolumn{4}{l}{\it Novel Dataset BMW}\\
Full Version             & BMW       & 72.28        & 74.26             \\
Adapter Version          & BMW       & \textbf{74.26}        & 74.26             \\
Continue on Full Version & BMW       & \textbf{74.26}        & --                 \\
Adapter on Full Version  & BMW       & 72.28        & --                 \\
New Full Version         & BMW       & 70.30        & --                 \\
New Adapter Version      & BMW       & 71.29        & --                 \\
\bottomrule
\end{tabular}}
\caption{Model 2 (MSTP Selection) Accuracy}
\label{tab:model2_results}
\end{table}

\begin{figure*}[t]
  \centering
  \begin{subfigure}[b]{0.33\textwidth}
    \includegraphics[width=\textwidth]{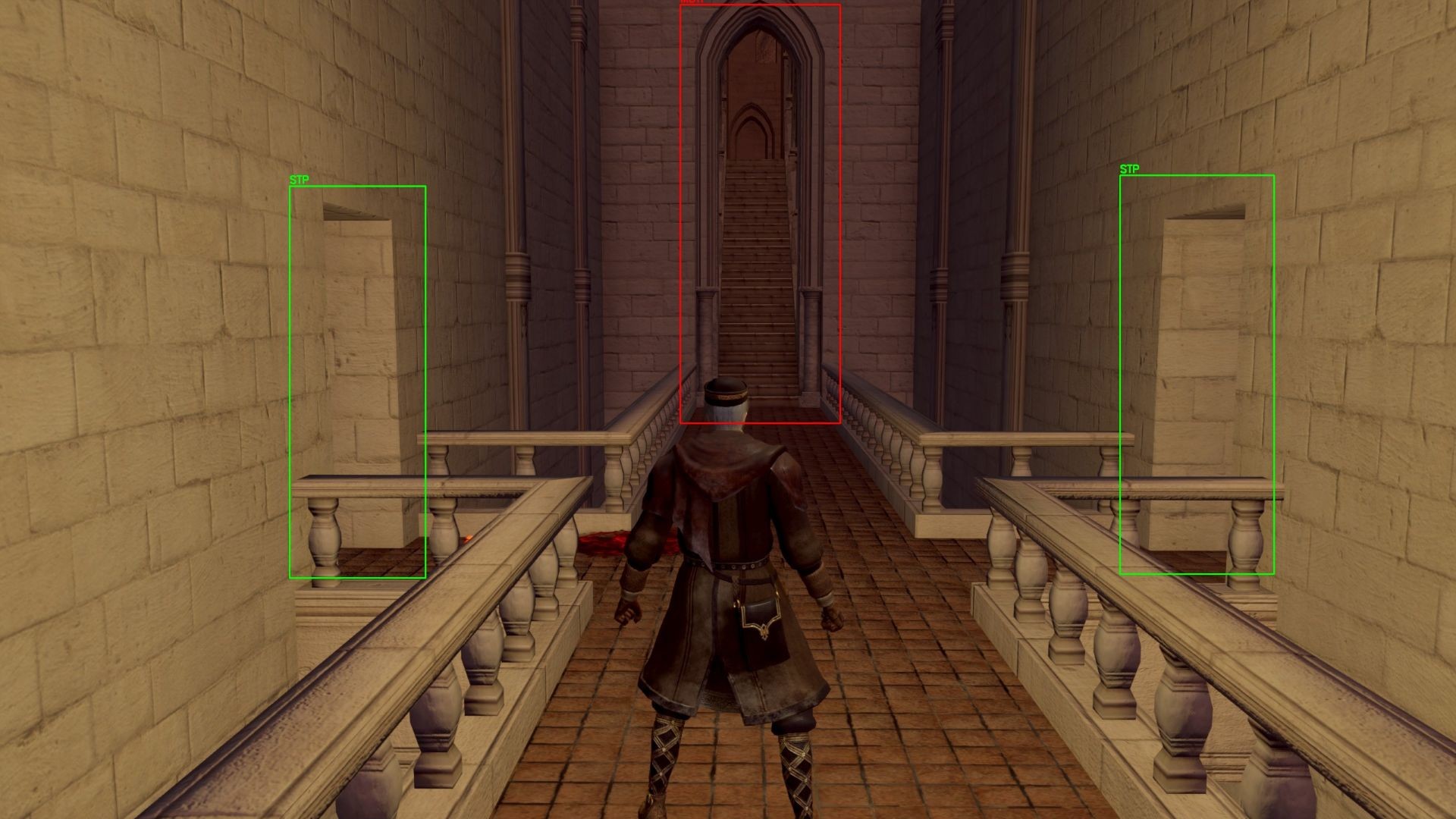}
    \caption{DS I (good): in a symmetric corridor, the doorway with the upward staircase breaks the symmetry, picked as the MSTP.}
  \end{subfigure}
  \hfill
  \begin{subfigure}[b]{0.33\textwidth}
    \includegraphics[width=\textwidth]{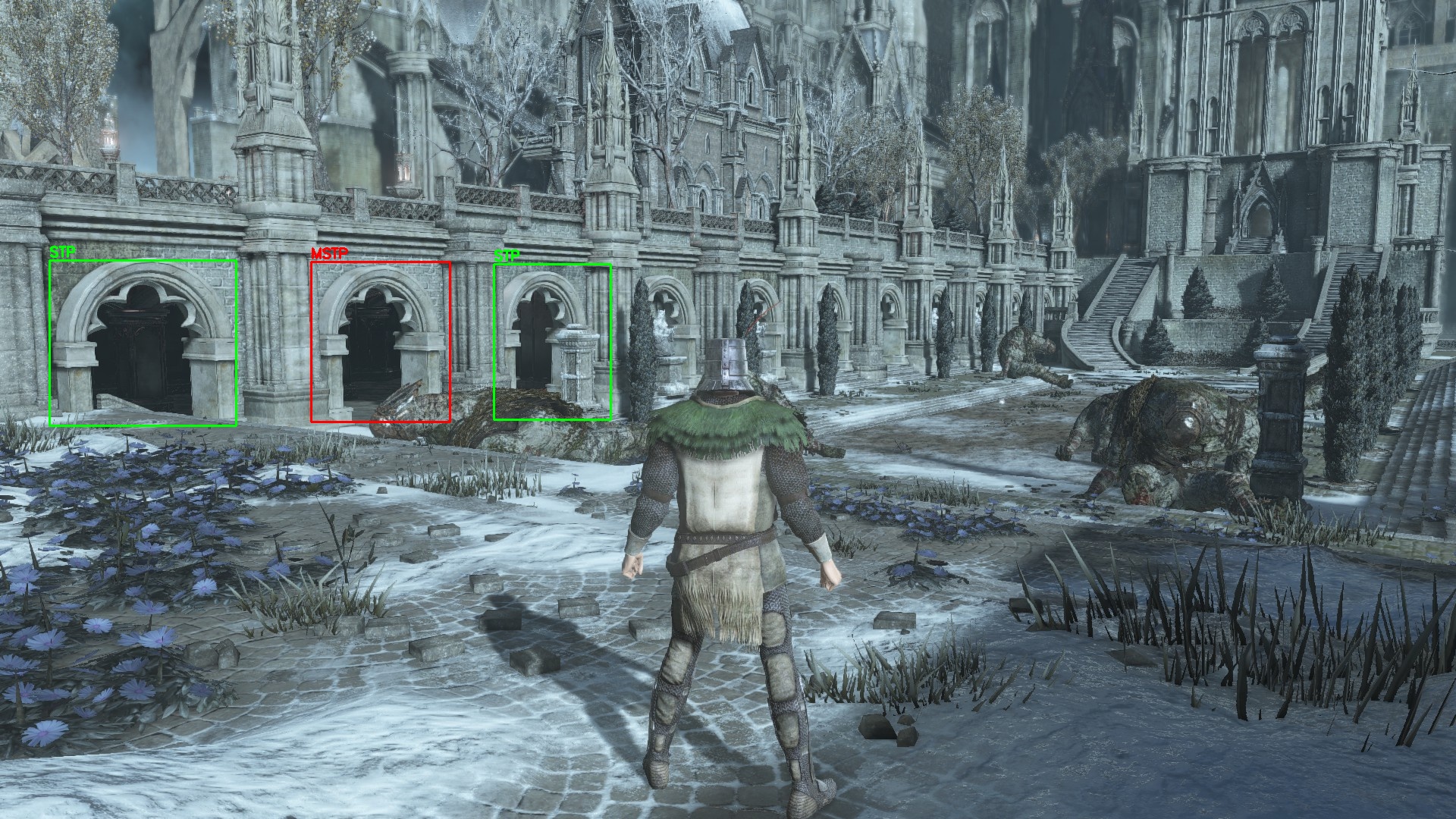}
    \caption{DS III (bad): repeated arched openings produce multiple high‐confidence STP candidates, while the real route is the far staircase.}
  \end{subfigure}
    \hfill
  \begin{subfigure}[b]{0.33\textwidth}
    \includegraphics[width=\textwidth]{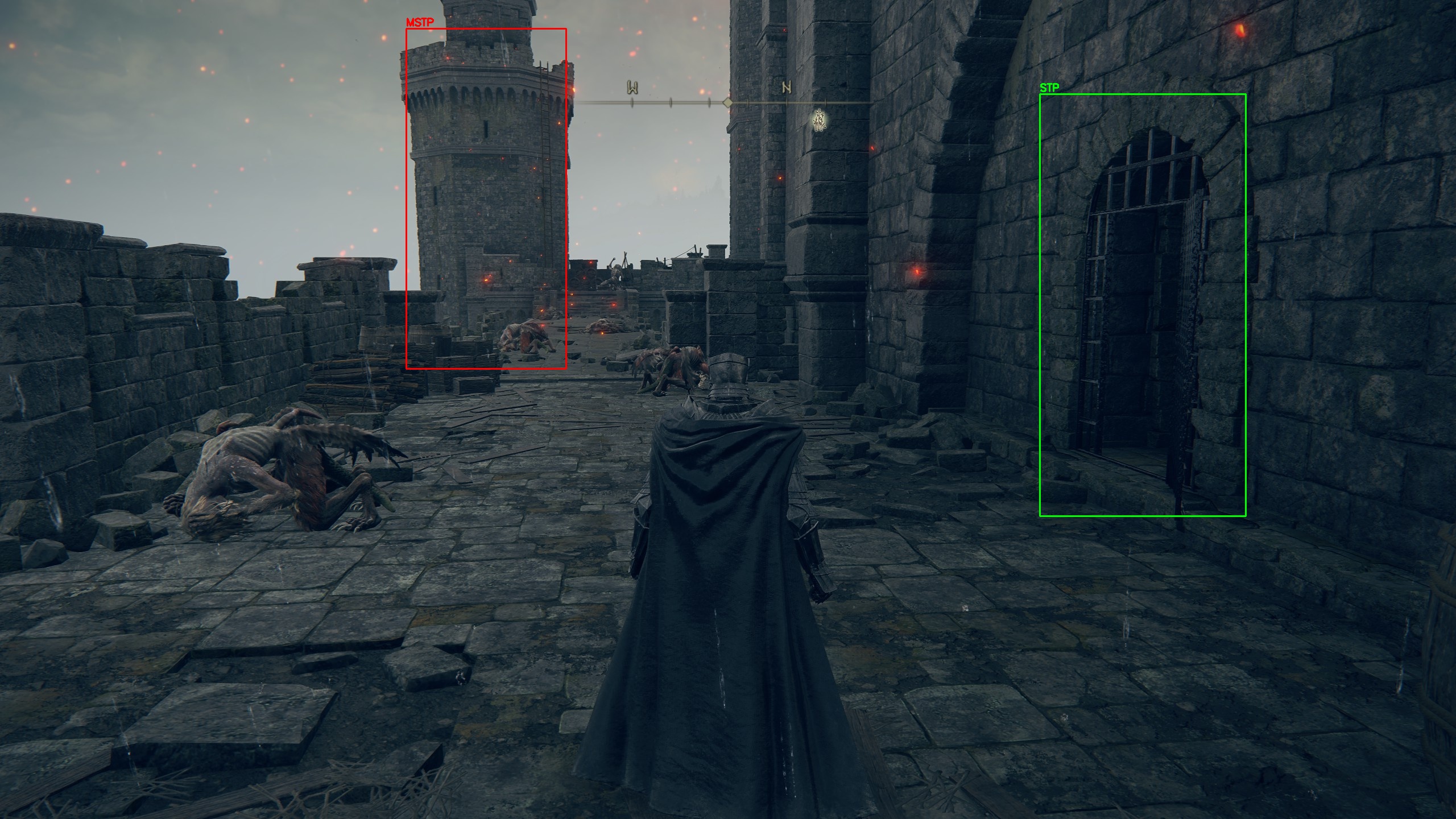}
    \caption{Elden Ring (good): the tower’s strong contrast against the background of sky leads to a high score, thus one clear MSTP.}
  \end{subfigure}
  
    \vspace{1ex}
    
  \begin{subfigure}[b]{0.33\textwidth}
    \includegraphics[width=\textwidth]{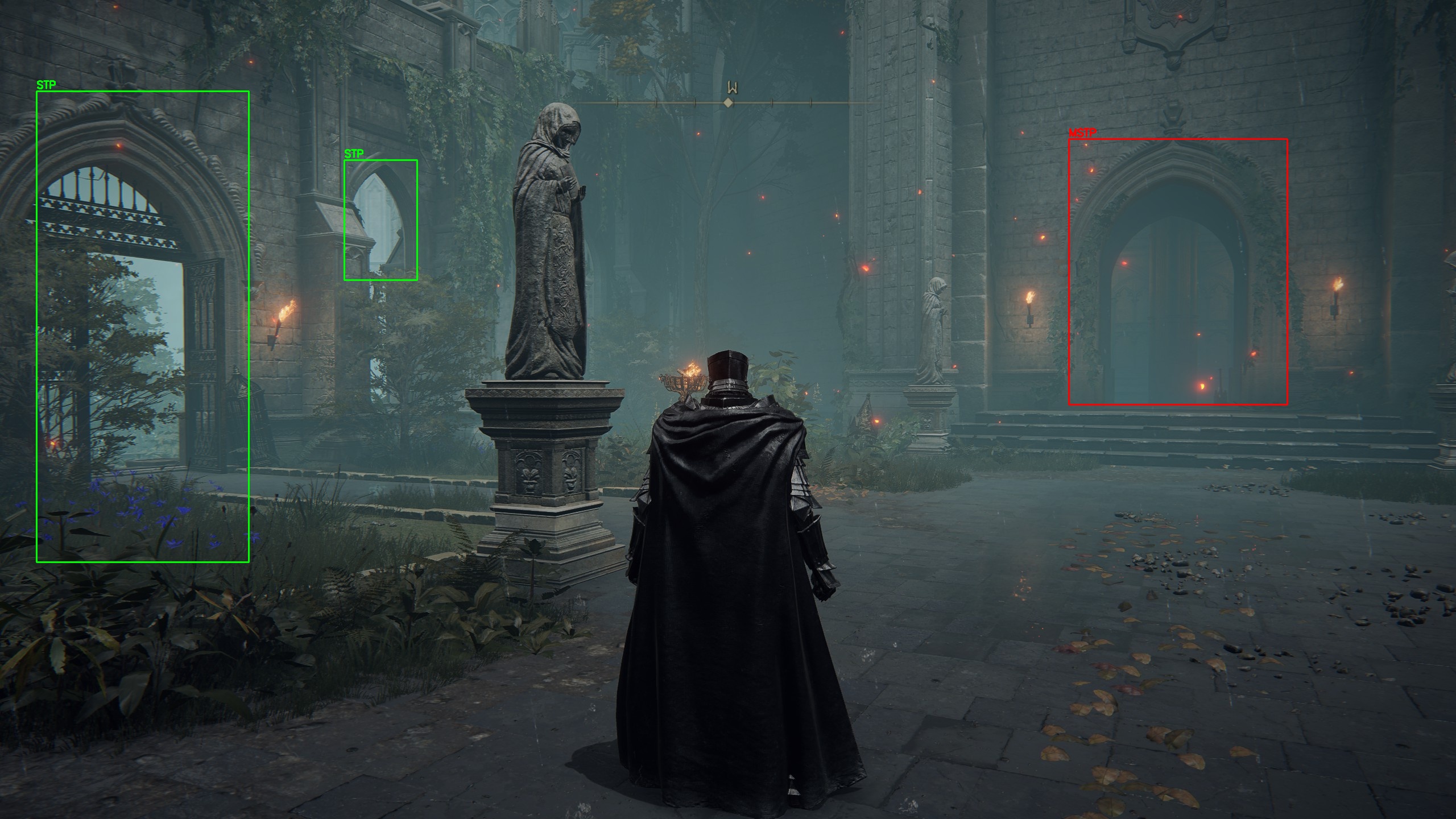}
    \caption{Elden Ring (bad): the model also marks a small, high-contrast window in the wall as an STP, though the correct MSTP is identified.}
  \end{subfigure}
  \hfill
  \begin{subfigure}[b]{0.33\textwidth}
    \includegraphics[width=\textwidth]{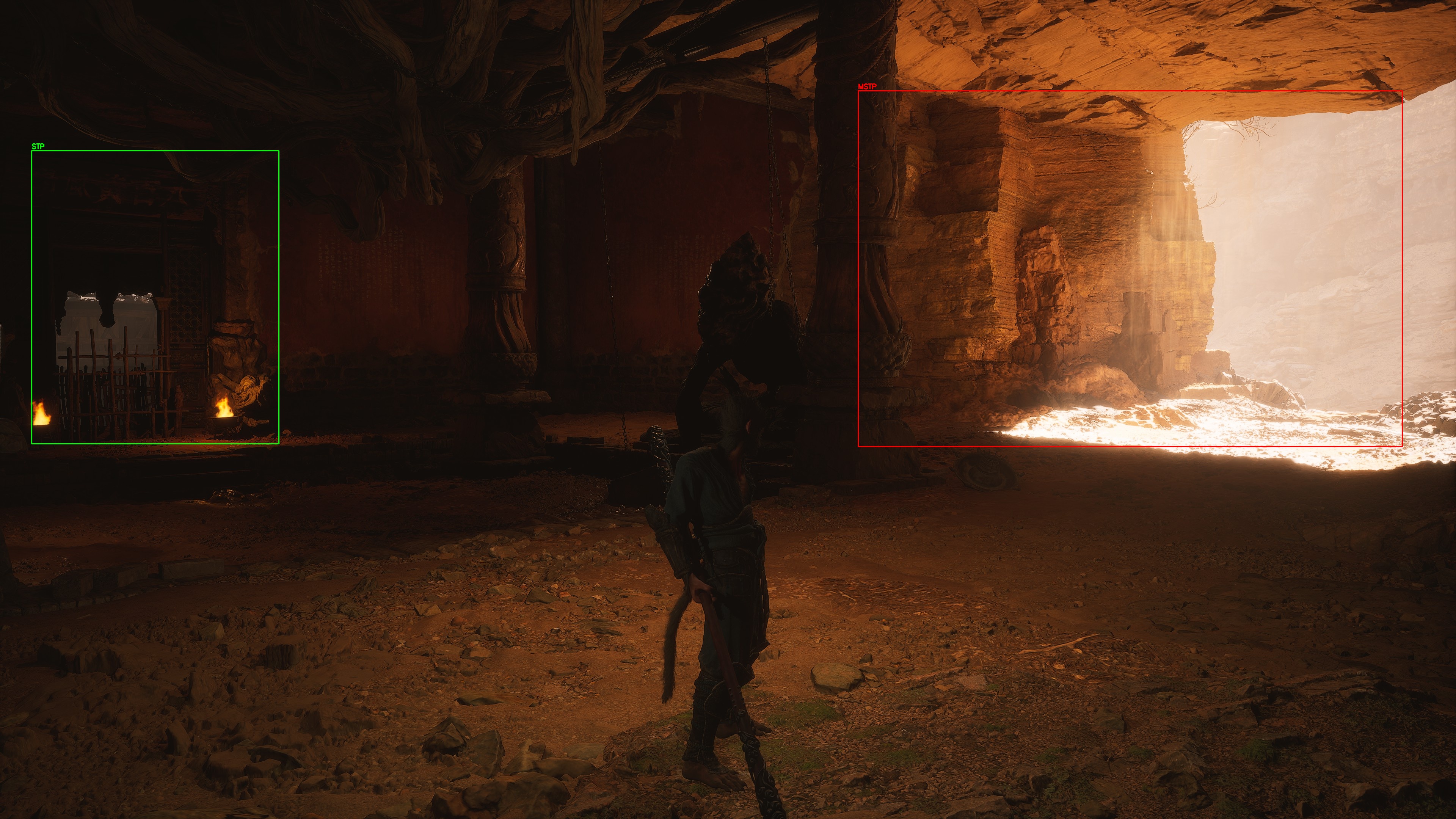}
    \caption{Black Myth (good): the bright canyon exit MSTP clearly beats another low contrast STP candidate.}
  \end{subfigure}
  \hfill
  \begin{subfigure}[b]{0.33\textwidth}
    \includegraphics[width=\textwidth]{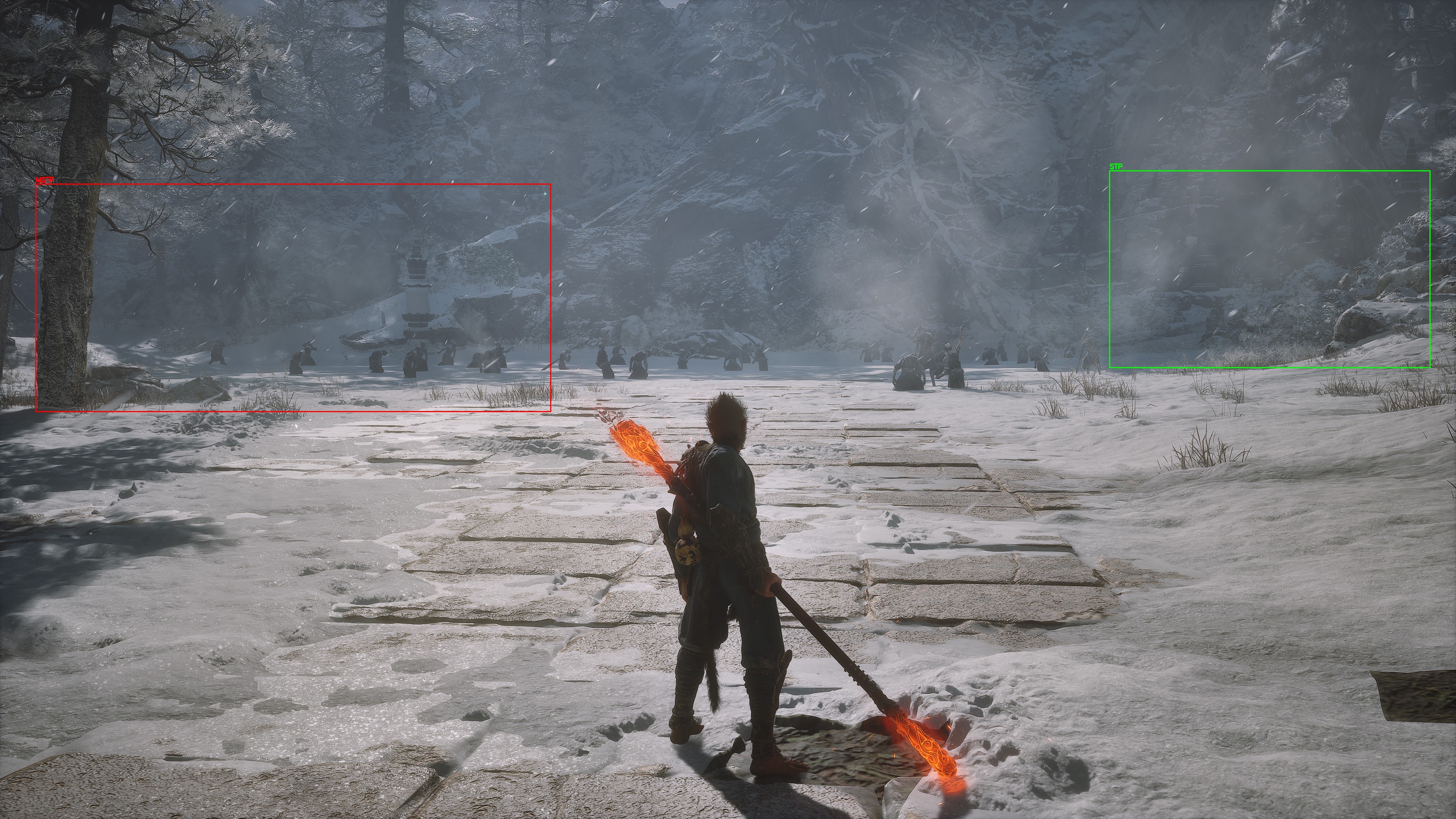}
    \caption{Black Myth (bad): the left cliff resembles a ramp, misleading the model; the true path is a plain gap behind the rocks on the right.}
  \end{subfigure}

  \caption{Qualitative outputs across four ARPGs.  
  Green boxes mark ground-truth STPs; red boxes show the model’s chosen MSTP.  
  Panels (a, c, e) succeed with clear local cues.  
  Panels (b) false high-contrast openings;  
  (d) window misidentified as STP due to misleading perspective; 
  (f) cliff face misread as ramp.}
  \label{fig:qualitative}
\end{figure*}

On the original Souls dataset, the Adapter‐only variant achieves 80.00\% accuracy—slightly above the Full model’s 77.89\%.  Retrieval fusion raises the Full model’s accuracy to 80.00\%, perfectly matching the Adapter baseline, while it has no effect on the Adapter‐only variant.  This confirms that for the relatively simple MSTP classification task, lightweight adapters already capture the necessary features, and semantic retrieval cues are largely redundant. We also evaluated the MSTP selection model using five random seeds. We compare against two simple baselines: (i) Random-1/k, which uniformly selects among the k detector proposals in a frame, and (ii) Center-most, which chooses the candidate whose box center is closest to the image center. Our method achieved a mean accuracy of 76.62\% ± 2.96\% (95\% CI [73.24\%, 80.00\%]), outperforming the strongest naive baseline (Center-most, 70.52\%). A paired McNemar’s test yielded p = 0.04, indicating that the improvement is statistically significant at the 5\% level. The relatively low variance across seeds further demonstrates the robustness and reliability of our approach.

Under extreme scarcity in \textbf{DS II} (20\% train), the adapter‐only model again leads with 73.75\% accuracy, compared to 66.25\% for the zero‐shot Full model. Retrieval fusion further boosts the Full model to 72.50\%, nearly closing the gap.  Continuing full fine‐tuning (73.75\%) matches the Adapter baseline, whereas scratch‐trained models (New Full: 72.50\%; New Adapter: 75.00\%) remain competitive but do not surpass the adapter‐only strategy.  These results underscore the robustness of adapter‐based transfer under low‐data conditions, and show that RAF can partially offset a lack of end‐to‐end fine‐tuning.

On the entirely new \textbf{BMW} domain, the same ordering persists: the Adapter‐only model reaches the highest accuracy (74.26\%), and retrieval fusion further elevates the Full model to 74.26\% (from its zero‐shot 72.28\%).  Full→Full on BMW also achieves 74.26\%, showing that full‐network fine‐tuning plus RAF yields parity with adapter‐only tuning.  Scratch‐trained baselines (New Full: 70.30\%, New Adapter: 71.29\%) lag behind.  Considering the general task difficulty, less than half the frames (\(47.2\%\)) contain only one STP; when restricted to multi-candidate cases (\(k\ge 2\)), all models stay above 50\% accuracy, with the adapter-only model around 60\%. In sum, adapter‐only fine‐tuning generalizes as well as—or better than—full‐network training across both Souls and non-Souls ARPGs, with RAF providing a consistent but modest uplift to the Full models.

\subsection{Real-Time Navigation Pilot Study} 

To verify that our offline metrics translate into usable in-game assistance, we conducted a small-scale pilot study involving live gameplay. A single player (an author) initiated each run at a level start point. At every navigational fork, the player followed the MSTP suggested by the \textbf{Full Version} model (trained on the Original Dataset with RAF enabled), using an inference threshold of 0.5 for STP detection to ensure a reasonable number of candidates for MSTP selection. The player only manually overrode the system's guidance if the chosen path led to an unambiguous dead-end, forcing a backtrack. We logged the total number of decision points encountered and the instances of such human intervention.

Across five representative levels—\textit{High Wall of Lothric, Catacombs of Carthus, Grand Archives} (from Dark Souls III); \textit{Forest of Fallen Giants} (from Dark Souls II); and \textit{Black Wind Mountain} (from Black Myth: Wukong)—the system made 63 navigational decisions, requiring manual correction 11 times, yielding an approximate success rate of 82.5\%. The per-level intervention/decision counts were: High Wall 2/15, Catacombs 0/8, Grand Archives 2/13, Forest of Fallen Giants 3/10, and Black Wind Mountain 4/17. On a single NVIDIA RTX 4090 GPU, the MSTP selector achieved a median inference latency of less than 20 ms. This performance is well within our decision-making interval of 0.2 seconds (200 ms) used during the live play (game running at 60 FPS), confirming real-time feasibility.

\textbf{Key Observations from Pilot Study:}

(1) \emph{Opposing View Choices:} The system, analyzing single frames, sometimes struggled when competing STPs were in entirely opposite camera directions (e.g., the treasure-mimic alcove versus the rooftop ladder in High Wall). It tended to favor the currently visible or more centrally framed option. We plan to investigate multi-view stitching or short temporal window analysis to address this.

(2) \emph{Potential Upward-Bias Artefact:} The model occasionally over-prioritized upward paths, particularly staircases (e.g., preferring a small upper stair before the Boss in High Wall of Lothric over the main downward progression stairs). This might stem from a prevalence of upward-leading MSTPs in the training data. Augmenting the dataset should help mitigate this.

(3) \emph{Near-Field Ambiguity in Dense Layouts:} In areas with dense clusters of visually similar STPs in close proximity, such as the numerous short stair flights in the Grand Archives, the model found fine-grained ranking less reliable. This suggests a need for training data with richer examples of such close-range, complex decision points.

(4) \emph{Promising Discovery of Unseen Paths:} Despite the challenges, the system showed encouraging generalization by identifying some designer-intended shortcuts or hidden paths not explicitly present in its training data. Notable examples include a concealed coffin passage in the Catacombs of Carthus and a hidden cave entrance (potentially leading to an optional boss) on Black Wind Mountain in BMW, even when the model was not trained on BMW data.

\subsection{Qualitative Results}
\label{sec:qualitative}

To better understand where our pipeline succeeds and where it fails, Figure~\ref{fig:qualitative} shows ``good'' and ``bad'' examples from our games.  In each pair, green boxes are detected STPs, red boxes the chosen MSTP.  
Overall, our model proves highly sensitive to local contrast and shape cues—quick to pick out crisp, high-contrast details—but this strength also leads it to over-rely on individual structural features. As a result, incidental windows, decorative arches, or rock textures with strong edge definitions are sometimes treated as intentionally designed transition points. This behavior is reasonable and difficult to avoid given our frame-only, single-region focus, and it remains valuable in photorealistic settings (e.g., BMW) where fine visual differences matter. Mitigating this over-dependence on isolated local cues will be a central challenge for future improvements.

\section{Limitations}
\label{sec:limitations}

Our work has several constraints. The dataset is modest in overall frame count and game title/genre representation, which may affect broader generalization. Our definitions of STP and MSTP are best suited for games with clearly defined main progression paths and can be subjective in more open-ended scenarios. The proposed baseline pipeline has not incorporated more specialized designs for explicit visual reasoning. We also have not evaluated the interpretability of our model’s decisions or provided mechanisms to explain its outputs. Furthermore, our experimental validation has primarily focused on static image analysis, complemented by a small-scale pilot study for real-time interaction; we have not conducted large-scale player trials to assess how well our model guides a diverse range of users in practice, which remains necessary for comprehensive evaluation.

Future work will directly address these limitations. We plan to expand the dataset in both volume and variety with community contributions, including more game genres and visual styles. A key direction is to evolve the concept of visual navigation beyond fixed STPs/MSTPs in linear paths, exploring models that can understand more dynamic navigational affordances in open-world or systemic game designs. Methodologically, we will investigate advanced architectures for improved visual understanding and temporal information processing from video data. We also intend to develop interpretability analyses for the rationale behind our model’s predictions. Large-scale user studies involving real-time interaction will be crucial for validating and refining its practical utility as both a player assistant and a design aid.

\section{Conclusion}
\label{sec:conclusion}

This paper addressed the challenge of automatically identifying designer-intended visual navigation cues in complex 3D RPG environments, formalizing the problem of recognizing STPs and the critical MSTPs from visual data. We introduced a two-stage deep learning pipeline that successfully detects potential STPs and subsequently identifies the MSTP using a combination of object detection, feature fusion, parameter-efficient adapters, and optional retrieval augmentation. Our aim was not to achieve state-of-the-art on a pre-existing benchmark, but rather to define this novel problem space and demonstrate its tractability.

The initial experiments, conducted on a diverse, newly created dataset from five Action RPGs, primarily served to establish the initial feasibility of our proposed solution and to highlight key performance characteristics. These tests revealed an important trade-off: for the dense STP-detection task, full network training is effective with enough data, while parameter-efficient adapter-only transfer is more robust for data-scarce scenarios and for MSTP selection across varying data regimes. Retrieval-augmented fusion showed potential as a low-cost enhancement. A real-time pilot further confirmed in-game feasibility. The insights gained from these initial evaluations affirm that the automated recognition of such visual cues is a valuable and solvable challenge.

In summary, by defining a new problem in automated visual navigation analysis and providing an empirically validated solution, this research opens promising avenues for creating AI systems that can more deeply comprehend and interact with the visual language of game environments, ultimately enhancing both player engagement and the art and efficiency of game design.

\bibliography{aaai25}

\begin{thebibliography}{28}
\providecommand{\natexlab}[1]{#1}

\bibitem[{Betsas et~al.(2025)Betsas, Georgopoulos, Doulamis, and Grussenmeyer}]{betsas2025deep}
Betsas, T.; Georgopoulos, A.; Doulamis, A.; and Grussenmeyer, P. 2025.
\newblock Deep Learning on 3D Semantic Segmentation: A Detailed Review.
\newblock \emph{Remote Sensing}, 17(2): 298.

\bibitem[{B{\o}e(2024)}]{boe2024role}
B{\o}e, R.~J. 2024.
\newblock \emph{The Role of Guidance Techniques on the Player Experience in Virtual Reality Games}.
\newblock Master's thesis, The University of Bergen.

\bibitem[{Csurka(2017)}]{csurka2017domain}
Csurka, G. 2017.
\newblock Domain adaptation for visual applications: A comprehensive survey.
\newblock \emph{arXiv preprint arXiv:1702.05374}.

\bibitem[{Dillman et~al.(2018)Dillman, Mok, Tang, Oehlberg, and Mitchell}]{dillman2018visual}
Dillman, K.~R.; Mok, T. T.~H.; Tang, A.; Oehlberg, L.; and Mitchell, A. 2018.
\newblock A visual interaction cue framework from video game environments for augmented reality.
\newblock In \emph{Proceedings of the 2018 CHI conference on human factors in computing systems}, 1--12.

\bibitem[{Fern{\'a}ndez-Vara(2011)}]{fernandez2011game}
Fern{\'a}ndez-Vara, C. 2011.
\newblock Game spaces speak volumes: Indexical storytelling.
\newblock \emph{Proceedings of DiGRA 2011 Conference}.

\bibitem[{Filén and Gemal(2024)}]{filen2024auditory}
Filén, P.; and Gemal, A. 2024.
\newblock {Auditory and Visual Feedback: Impact on Player Performance and Comprehension in First Person Shooters}.
\newblock Bachelor’s thesis, KTH Royal Institute of Technology.

\bibitem[{Gibson(2014)}]{gibson2014ecological}
Gibson, J.~J. 2014.
\newblock \emph{The ecological approach to visual perception: classic edition}.
\newblock Psychology press.

\bibitem[{Hare and Tang(2022)}]{hare2022player}
Hare, R.; and Tang, Y. 2022.
\newblock Player modeling and adaptation methods within adaptive serious games.
\newblock \emph{IEEE Transactions on Computational Social Systems}, 10(4): 1939--1950.

\bibitem[{Houlsby et~al.(2019)Houlsby, Giurgiu, Jastrzebski, Morrone, De~Laroussilhe, Gesmundo, Attariyan, and Gelly}]{houlsby2019parameter}
Houlsby, N.; Giurgiu, A.; Jastrzebski, S.; Morrone, B.; De~Laroussilhe, Q.; Gesmundo, A.; Attariyan, M.; and Gelly, S. 2019.
\newblock Parameter-efficient transfer learning for NLP.
\newblock In \emph{International conference on machine learning}, 2790--2799. PMLR.

\bibitem[{Hyde-Smith(2023)}]{hyde2023using}
Hyde-Smith, P. 2023.
\newblock \emph{Using Object Detection to Navigate a Game Playfield}.
\newblock Master's thesis, Marquette University.

\bibitem[{{Interaction Design Foundation (IxDF)}(2023)}]{ixdf_games_user_research_qualitative}
{Interaction Design Foundation (IxDF)}. 2023.
\newblock What is Games User Research?

\bibitem[{Irshad, Perkis, and Azam(2021)}]{irshad2021wayfinding}
Irshad, S.; Perkis, A.; and Azam, W. 2021.
\newblock Wayfinding in virtual reality serious game: An exploratory study in the context of user perceived experiences.
\newblock \emph{Applied Sciences}, 11(17): 7822.

\bibitem[{Jung, Yang, and Min(2021)}]{jung2021improving}
Jung, M.; Yang, H.; and Min, K. 2021.
\newblock Improving deep object detection algorithms for game scenes.
\newblock \emph{Electronics}, 10(20): 2527.

\bibitem[{Lin et~al.(2024)Lin, Li, Chen, and Xiong}]{lin2024design}
Lin, X.; Li, R.; Chen, Z.; and Xiong, J. 2024.
\newblock {Design Strategies for VR Science and Education Games from an Embodied Cognition Perspective: A Literature-Based Meta-Analysis}.
\newblock \emph{Frontiers in Psychology}, 14: 1292110.

\bibitem[{Liu et~al.(2021)Liu, Snodgrass, Khalifa, Risi, Yannakakis, and Togelius}]{liu2021deep}
Liu, J.; Snodgrass, S.; Khalifa, A.; Risi, S.; Yannakakis, G.~N.; and Togelius, J. 2021.
\newblock Deep learning for procedural content generation.
\newblock \emph{Neural Computing and Applications}, 33(1): 19--37.

\bibitem[{Lynch(2023)}]{lynch2023image}
Lynch, K. 2023.
\newblock The Image of the City (1960).
\newblock In \emph{Anthologie zum Städtebau. Band III: Vom Wiederaufbau nach dem Zweiten Weltkrieg bis zur zeitgenössischen Stadt}, 481--488. Gebr. Mann Verlag.

\bibitem[{Marcus et~al.(2025)Marcus, Paay, Langenheim, and Yang}]{marcus2025hci}
Marcus, W.; Paay, J.; Langenheim, N.; and Yang, T. 2025.
\newblock {HCI Methods Supporting Urban Design Evaluation Using Virtual Environments}.
\newblock \emph{Interacting with Computers}, iwaf014.

\bibitem[{Mehta(2025)}]{mehta2025role}
Mehta, N. 2025.
\newblock The Role of AI in Game Development and Player Experience.
\newblock \emph{Available at SSRN 5101269}.

\bibitem[{Miller(2024)}]{miller2024automated}
Miller, D. 2024.
\newblock Automated Playtest and Crash Analyzer (APCA): An AI that Playtests Video Games to Detect Bugs and Potential Crashes.
\newblock \emph{ScienceOpen Posters}.

\bibitem[{Naying et~al.(2023)Naying, Yuexian, Khalid, and Iida}]{naying2023computational}
Naying, G.; Yuexian, G.; Khalid, M. N.~A.; and Iida, H. 2023.
\newblock A computational game experience analysis via game refinement theory.
\newblock \emph{Telematics and Informatics Reports}, 9: 100039.

\bibitem[{Omidshafiei et~al.(2020)Omidshafiei, Tuyls, Czarnecki, Santos, Rowland, Connor, Hennes, Muller, P{\'e}rolat, Vylder et~al.}]{omidshafiei2020navigating}
Omidshafiei, S.; Tuyls, K.; Czarnecki, W.~M.; Santos, F.~C.; Rowland, M.; Connor, J.; Hennes, D.; Muller, P.; P{\'e}rolat, J.; Vylder, B.~D.; et~al. 2020.
\newblock Navigating the landscape of multiplayer games.
\newblock \emph{Nature communications}, 11(1): 5603.

\bibitem[{Patel et~al.(2015)Patel, Gopalan, Li, and Chellappa}]{patel2015visual}
Patel, V.~M.; Gopalan, R.; Li, R.; and Chellappa, R. 2015.
\newblock Visual domain adaptation: A survey of recent advances.
\newblock \emph{IEEE signal processing magazine}, 32(3): 53--69.

\bibitem[{Qi and Li(2023)}]{qi2023application}
Qi, J.; and Li, H. 2023.
\newblock Application of a semantic segmentation method based on deep learning in unity scene construction.
\newblock In \emph{Third International Conference on Computer Vision and Pattern Analysis (ICCPA 2023)}, volume 12754, 774--777. SPIE.

\bibitem[{Ren et~al.(2015)Ren, He, Girshick, and Sun}]{ren2015faster}
Ren, S.; He, K.; Girshick, R.; and Sun, J. 2015.
\newblock Faster r-cnn: Towards real-time object detection with region proposal networks.
\newblock \emph{Advances in neural information processing systems}, 28.

\bibitem[{Xu and Verbrugge(2025)}]{11114135}
Xu, K.; and Verbrugge, C. 2025.
\newblock Quantitative Analysis of Visual Guidance in Level Transitions Using Multimodal Visual Metrics.
\newblock In \emph{2025 IEEE Conference on Games (CoG)}, 1--8.

\bibitem[{Yalçınkaya-Doma(2024)}]{yalcinakaya_material_matters_2024}
Yalçınkaya-Doma, G. 2024.
\newblock \emph{Material Matters: The Effects of Materials On Spatial Experience and Navigation in Video Games}.
\newblock Master's thesis, Bahçeşehir University.

\bibitem[{Yannakakis and Togelius(2018)}]{yannakakis2018artificial}
Yannakakis, G.~N.; and Togelius, J. 2018.
\newblock \emph{Artificial intelligence and games}, volume~2.
\newblock Springer.

\bibitem[{Yazdani et~al.(2025)Yazdani, Bosaghzadeh, Ebrahimpour, and Dornaika}]{yazdani2025computational}
Yazdani, H.; Bosaghzadeh, A.; Ebrahimpour, R.; and Dornaika, F. 2025.
\newblock A Computational--Cognitive Model of Audio-Visual Attention in Dynamic Environments.
\newblock \emph{Big Data and Cognitive Computing}.

\end{thebibliography}

\end{document}